# SynSeg-Net: Synthetic Segmentation Without Target Modality Ground Truth

Yuankai Huo*, Zhoubing Xu, Hyeonsoo Moon, Shunxing Bao, Albert Assad, Tamara K. Moyo, Michael R. Savona, Richard G. Abramson, and Bennett A. Landman

*Abstract*— A key limitation of deep convolutional neural networks (DCNN) based image segmentation methods is the lack of generalizability. Manually traced training images are typically required when segmenting organs in a new imaging modality or from distinct disease cohort. The manual efforts can be alleviated if the manually traced images in one imaging modality (e.g., MRI) are able to train a segmentation network for another imaging modality (e.g., CT). In this paper, we propose an end-to-end synthetic segmentation network (SynSeg-Net) to train a segmentation network for a target imaging modality without having manual labels. SynSeg-Net is trained by using (1) unpaired intensity images from source and target modalities, and (2) manual labels only from source modality. SynSeg-Net is enabled by the recent advances of cycle generative adversarial networks (CycleGAN) and DCNN. We evaluate the performance of the SynSeg-Net on two experiments: (1) MRI to CT splenomegaly synthetic segmentation for abdominal images, and (2) CT to MRI total intracranial volume synthetic segmentation (TICV) for brain images. The proposed end-to-end approach achieved superior performance to two stage methods. Moreover, the SynSeg-Net achieved comparable performance to the traditional segmentation network using target modality labels in certain scenarios. The source code of SynSeg-Net is publicly available [2].

*Index Terms*— Synthesis, Segmentation, Splenomegaly, TICV, Synthetic Segmentation, GAN, Adversarial, DCNN, Convolutional

## I. INTRODUCTION

Deep learning techniques have proven effective for medical image synthesis across (1) different sequencing types within the same image modality (e.g., between T1w, T2w, PD, FLAIR etc.) and (2) different imaging modalities (e.g. MRI to CT, CT to MRI etc.) [1]. While there are impediments to use the synthetic images directly in clinical practice, synthetic images have been shown to be an effective intermediate representation for image processing including registration [2], data augmentation [3], and segmentation [4]. Historically, paired training data for both imaging modalities were typically required for image synthesis. Recent advances with cycle generative adversarial networks (CycleGAN) [5] have demonstrated high quality cross-modality image synthesis without paired data.

In this paper, we propose an end-to-end synthetic segmentation network (SynSeg-Net) to train a DCNN segmentation network without having manual labels on the target imaging modality. The network is trained by unpaired source and target modality images with manual segmentations only on the source modality (Figure 1). This method alleviates the manual segmentation efforts for the medical image analyses by taking the advantage of cross-modality image synthesis learning.

To evaluate the segmentation performance of the proposed SynSeg-Net, two experiments were employed. The first

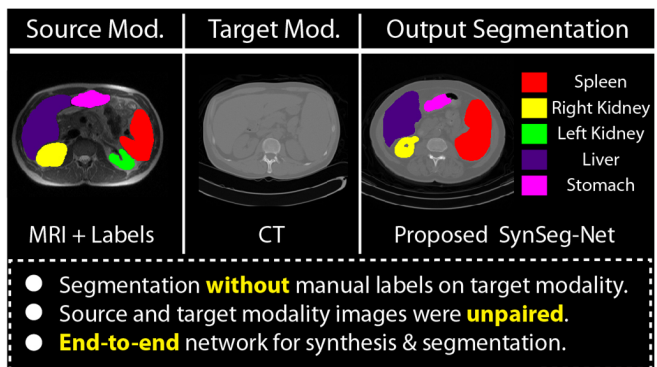

Fig. 1. The proposed synthetic segmentation network (SynSeg-Net) is able to train a CT splenomegaly segmentation network from unpaired MRI and CT training images without using manual CT labels.

This research was supported by NSF CAREER 1452485 (Landman), NIH grants 5R21EY024036 (Landman), R01EB017230 (Landman), 1R21NS064534 (Prince/Landman), 1R01NS070906 (Pham), 2R01EB006136 (Dawant), 1R03EB012461 (Landman), NCI Cancer Center Support Grant (P30 CA068485), and R01NS095291 (Dawant). This research was also supported by the Vanderbilt-Incyte Research Alliance Grant (Savona/ Abramson/ Landman). This research was also supported by InCyte Corporation (Abramson/Landman). This research was conducted with the support from Intramural Research Program, National Institute on Aging, NIH. This study was also supported by NIH 5R01NS056307, 5R21NS082891 and in part using the resources of the Advanced Computing Center for Research and Education (ACCRE) at Vanderbilt University, Nashville, TN. This project was supported in part by ViSE/VICTR VR3029 and the National Center for Research Resources, Grant UL1 RR024975-01, and is now at the National Center for Advancing Translational Sciences, Grant 2 UL1 TR000445-06. The content is solely the responsibility of the authors and does not necessarily represent the official views of the NIH. We appreciate the NIH S10 Shared Instrumentation Grant 1S10OD020154-01 (Smith), Vanderbilt IDEAS grant (Holly-Bockelmann, Walker, Meiler, Palmeri, Weller), and ACCRE's Big Data TIPs grant from Vanderbilt University.

*Y. Huo is with the Department of Electrical Engineering and Computer Science, Vanderbilt University, Nashville, TN 37235 USA (e-mail: yuiankai.huo@vanderbilt.edu)

Z. Xu, H Moon, S. Bao, and B. A. Landman are with the Department of Electrical Engineering and Computer Science, Vanderbilt University, TN 37235 USA

T. K. Moyo and M. R. Savona is with the Department of Medicine, Vanderbilt University Medical Center. TN 37235 USA

A. Assad is with Incyte Corporation, Delaware 19803 USA

R. G. Abramson is with the Department of Radiology and Radiological Science, Vanderbilt University Medical Center. TN 37235 USA

[2] https://github.com/MASILab/SynSeg-Net

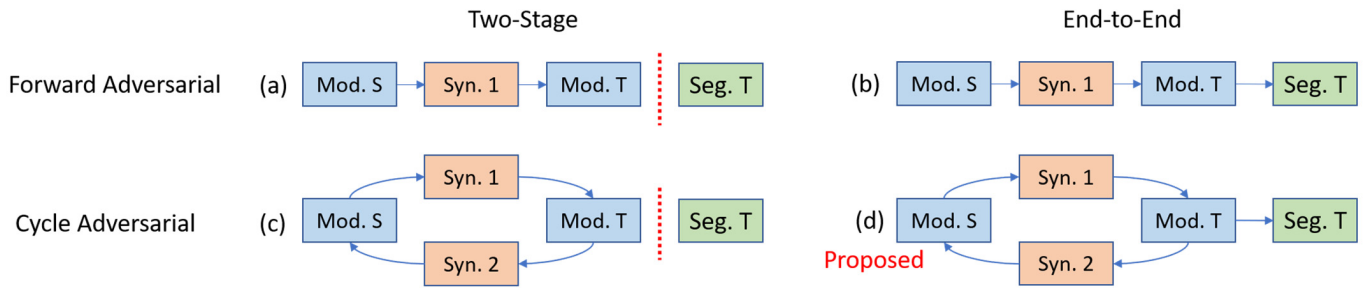

Fig. 2. This figure illustrates the prevalent strategies of performing segmentation without ground truth in the target modality. "Mod. S" means source modality images while "Mod. T" represents target modality images. "Syn. 1" is a source to target transformation generator, while "Syn. 2" is a target to source transformation generator. "Seg. T" is the segmentation network for target modality. (a) is the a two-stage framework that considered the synthesis (left side of the red dash line) and segmentation (right side of the red dash line) as two independent training stages. (b) connects the synthesis and segmentation network into an end-to-end fashion. (c) employs the latest CycleGAN framework as the synthesis network for unpaired cross-modality image synthesis (left side of the red dash line), and then performs another independent training stage for segmentation (right side of the red dash line). (d) is the proposed method which integrate the cycle adversarial synthesis and segmentation into a end-to-end framework.

experiment performed CT splenomegaly (extraordinary large spleen) synthetic segmentation without having any spleen labels on CT images. The second experiment performed MRI total intracranial volume (TICV) synthetic segmentation without having any TICV labels on MRI. From the empirical validations, the proposed end-to-end approach achieved superior performance to the two stage methods. Moreover, the SynSeg-Net achieved comparable performance to the traditional way of training a segmentation network using target modality labels in certain scenarios. Note that the "comparable performance" in this paper is defined as two methods do not show statistically significant differences on segmentation performance.

This work extends our previous conference paper [6] with the following new efforts: (1) the methodology is presented in greater detail, (2) new external validations (MRI to CT) were provided for CT splenomegaly synthetic segmentation, (3) total intracranial volume segmentation was provided as a new experiment (CT to MRI), and (4) the source code of SynSeg-Net has been made publicly available at https://github.com/MASILab/SynSeg-Net.

## II. RELATED WORKS

### A. Cross-modality Image Synthesis

Medical image synthesis is defined as the generation of realistic images through learning models [1]. From a technical perspective, image synthesis can be achieved from a generative model (e.g., from noise) or a cross-modality adaptation model (e.g., from MRI to CT). Our work is mostly related to the cross-modality image synthesis approaches, in which a synthetic image in target imaging modality is synthesized from a real image in source imaging modality.

Historically, cross-modality image synthesis methods can be ascribed to three categories (1) registration-based methods, (2) intensity-based methods, and (3) deep learning based methods. The registration-based cross-modality image synthesis methods were inspired by Millar et al. [2], in which the synthetic images were achieved by registering a subject image to a collection of co-registered images. Then, Burgos et al. [7] extended this idea to a multi-atlas information propagation scheme by integrating multi-atlas registration and intensity fusion, and applied on MRI to CT synthesis. Cardoso et al. [8] proposed a variant of this approach by introducing a multi-atlas generative model for image synthesis and outlier detection. The second family of the cross-modality image synthesis approaches is intensity-based methods, whose principle is to learn an intensity transformation function to map source intensities to target intensities [9-16].

Herein, we focus on the third family - deep learning based image synthesis methods. In [17], a location-sensitive deep synthesis method was introduced to utilize the both intensity and spatial information between modalities during training stage. Sevetlidis et al. proposed a deep encoder-decoder network [18] using a patch-based learning fashion. Xiang et al. proposed a deep embedding convolutional neural network [19], which utilize the intermediate feature maps between MRI and CT scans. Nie et al. [20] proposed a context-aware generative adversarial network to generate CT images from MRI images.

Recently, Goodfellow et al. proposed generative adversarial networks (GANs) [21] that provided a new perspective of image synthesis and domain adaptation in using either paired training images [22] or unpaired images [5]. GAN-based methods have been successfully applied to a variety of computer vision problems [23, 24] and have been adapted to medical imaging community [20, 25-27]. Compared with previous adversarial learning based synthesis method, the cycle consistent loops leads to more representative synthetic images.

### B. Synthetic Segmentation

One major application of image synthesis is to leverage segmentation performance. Iglesias et al. demonstrated that the synthesized MRI images could improve the segmentation performance (Figure 2a) [4]. Several studies used the adversarial learning as an extra GAN-based supervision on medical image segmentation networks [28-31]. In this study, we focus on the synthetic segmentation, which used the synthetic images as training images to train a segmentation network in target imaging modalities.

Figure 2 presents the different strategies for synthetic segmentation. Kamnitsas et al. [32] introduced unsupervised domain adaptation for brain lesion segmentation (Figure 2b). It reveals the possibility of training a lesion segmentation network using cross-modality synthetic segmentation. However, (1) the source imaging sequence (GE) and target imaging sequence (SWI) are still from to the same MRI modality. (2) Overlapped image modalities (e.g., FLAIR, T2, PD, MPRAGE) were used in both source and target imaging modalities to ensure performance. Cross-modality synthetic segmentation on two independent imaging mechanisms (e.g. MRI to CT) without having overlapped imaging modalities is appealing.

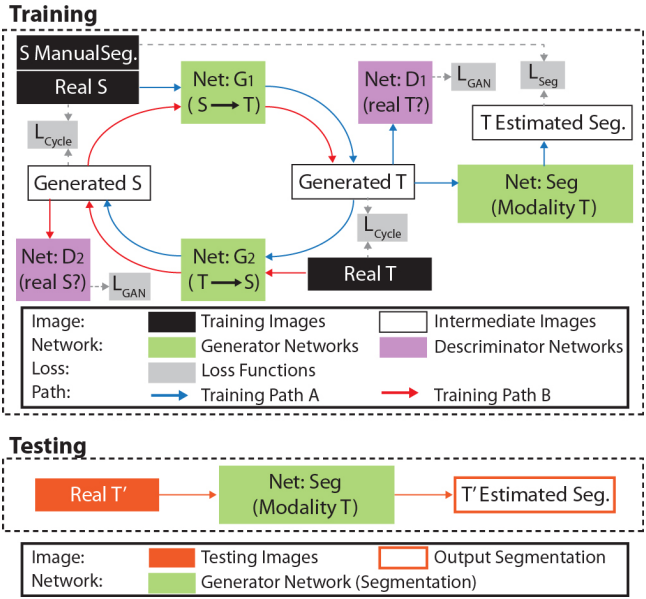

Fig. 3. The upper panel showed the network structure of the proposed SynSeg-Net during training stages. The left side was the CycleGAN synthesis subnet, where $S$ was MRI and $T$ was CT. $G_1$ and $G_2$ were the generators while $D_1$ and $D_2$ were discriminators. The right subnet was the segmentation subnet $Seg$ for an end-to-end training. Loss function were added to optimize the SynSeg-Net. The lower panel showed the network structure of SynSeg-Net during testing stage. Only the trained subnet $Seg$ was used to segment a testing image from target imaging modality.

Recently, the cycle generative adversarial networks (CycleGAN) [5] provided a promising tool for cross-modality synthesis from unpaired training images [33, 34]. With CycleGAN, one is able to synthesize the images for one imaging modality (e.g., MRI) while targeting another imaging modality (e.g., CT). Using CycleGAN, Chartsias et al. [35] proposed an CT to MRI synthesis method, and then trained another independent MRI segmentation network (called "Seg.") using the synthetic MRI images (Figure 2c). Although still using manual labels for both two modalities, this two-stage framework (we refer to as "CycleGAN+Seg.") revealed a promising direction of integrating cycle adversarial networks in synthetic segmentation.

Building upon CycleGAN, Zhang et al. [3] and our group [6] proposed end-to-end synthesis and segmentation networks. Zhang et al. [3] focus on leveraging both synthesis and segmentation performance simultaneously using both true images and manual labels on both MRI and CT. Therefore, the manual segmentation on target imaging modalities have still been used. By contrast, Huo et al. [6] introduced the end-to-end synthesis and segmentation network, which designed a synthetic segmentation network without using manual labels in target imaging modality. In this paper, we described such method with more detailed descriptions. Moreover, external validation and new experiments were employed to evaluate the proposed method as well as the baseline methods.

III. METHOD

Figure 3 introduces the network design, while preprocessing, postprocessing, hyperparameters and the experimental platforms are presented below.

A. Preprocessing

The intensities of every input MRI scan were normalized to 0-1 scale such that the highest 2.5% and lowest 2.5% intensities were excluded from the normalization to reduce the outliers' effects. For CT, the voxels whose HU values were greater than 1000 were set to 1000, whose HU values were less than -1000 were set to -1000. Then, the intensities between -1000 to 1000 were normalized to 0-1 scale. Next, the axial slices from normalized intensity image volume (both MRI and CT) were resampled to $256 \times 256$ using bilinear interpolation, while the corresponding segmentation axial slices were resampled to the same resolution using nearest neighbor interpolation. Hence, the same image dimensions ($256 \times 256$) were match for both modalities, following CycleGAN [5].

B. SynSeg-Net

Figure 3 presents the network structure of SynSet-Net, where "$S$" indicates the source imaging modality (e.g., MRI), while "$T$" indicates the target imaging modality (e.g., CT). The SynSeg-Net consisted of two major portions: cycle synthesis subnet and segmentation subnet.

1) Cycle Synthesis Subnet

The 9 block ResNet (defined in [5, 36]) was employed as the two generators $G_1$ and $G_2$. The generator $G_1$ transferred a real image $x$ in modality $S$ to a synthetic image $G_1(x)$ in modality $T$, while the generator $G_2$ synthesized a real image $y$ in modality $T$ to a synthetic image $G_2(y)$ in modality $S$. Next the PatchGAN (defined in [5, 37]) was used as the two adversarial discriminators $D_1$ and $D_2$. $D_1$ determined whether a provided image is a synthetic image $G_1(x)$ or a real image $y$, while $D_2$ judged whether a provided image is a synthetic image $G_2(y)$ or a real image $x$. When deploying such network on unpaired images from modality $S$ and $T$, two forward training paths (Path A and Path B) were used (in Figure 3).

2) Segmentation Subnet

Since the final aim of the proposed SynSeg-Net was to perform end-to-end synthetic segmentation, we concatenate a segmentation network "$Seg$" after $G_1$ directly, as an extension of the training Path A. To be consistent with the cycle synthesis subnet, the same the 9 block ResNet [5, 36] were used as $S$, whose network structure was identical to $G_1$.

3) Loss Functions

In SynSeg-Net, five loss functions have been used during the training stage. After discriminators $D_1$ and $D_2$, two adversarial loss functions were used to train the adversarial generators $G_1$ and $G_2$.

$$\mathcal{L}_{\text{GAN}}(G_1, D_1, S, T) = E_{y \sim T}[\log D_1(y)] \qquad (1)$$

$$+E_{x\sim S}[\log(1-D_1(G_1(x)))]$$
$$\mathcal{L}_{\text{GAN}}(G_2,D_2,T,S) = E_{x\sim S}[\log D_2(x)] \quad (2)$$
$$+E_{y\sim T}[\log(1-D_2(G_2(y)))]$$

Meanwhile, two cycle consistent loss functions were used to minimize the difference between true images and cycle reconstructed images.

$$\mathcal{L}_{\text{cycle}}(G_1,G_2,S) = E_{x\sim A}[\|G_2(G_1(x))-x\|_1] \quad (3)$$
$$\mathcal{L}_{\text{cycle}}(G_2,G_1,T) = E_{y\sim B}[\|G_1(G_2(y))-y\|_1] \quad (4)$$

The last loss function is the segmentation loss, which was the weighted cross entropy loss.

$$\mathcal{L}_{\text{seg}}(Seg,G_1,S) = -\sum_i m_i \log(Seg(G_1(x_i))) \quad (5)$$

After defining five loss functions, we added them together by assigning different weights.

$$\mathcal{L}_{\text{total}} = \lambda_1 \cdot L_{\text{GAN}}(G_1,D_1,S,T) + \lambda_2 \cdot \mathcal{L}_{\text{GAN}}(G_2,D_2,T,S)$$
$$+\lambda_3 \cdot \mathcal{L}_{\text{cycle}}(G_1,G_2,S) + \lambda_4 \cdot \mathcal{L}_{\text{cyle}}(G_2,G_1,T) \quad (6)$$
$$+\lambda_5 \cdot \mathcal{L}_{\text{seg}}(Seg,G_1,S)$$

### C. Training and Testing

In all experiments, the lambdas were empirically set to $\lambda_1 = 1$, $\lambda_2 = 1$, $\lambda_3 = 10$, $\lambda_4 = 10$, $\lambda_5 = 1$. The $\lambda_1$ to $\lambda_4$ were chosen using the same values in the original CycleGAN paper [5], where $\lambda_5$ was simply assigned it to 1 without tuning for different applications in this study. The Adam optimizer [5] was used to minimize the $\mathcal{L}_{\text{total}}$. The number of input and output channels of all networks are all one except $Seg$, which had seven output channels (background, spleen, left kidney, right kidney, stomach, liver, body mask). The Adam learning rate was 0.0001 for $G_1$, $G_2$ and $Seg$ and 0.0002 for $D_1$ and $D_2$.

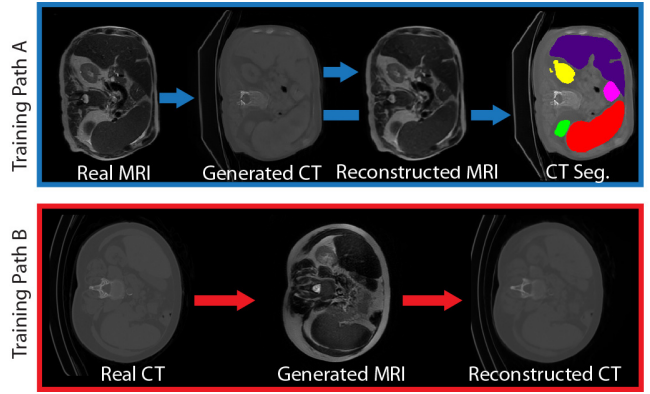

Fig. 4. The intermediate results of the real, synthesized, and reconstructed images as well as segmentations in training Path A and Path B.

In testing stage, only the segmentation network $Seg$ was employed by SynSeg-Net (Figure 3). To segment a testing scan in the target modality, the testing scan was normalized to 0-1. Next, the axial slices from normalized testing image volume were resampled to $256 \times 256$ using bilinear interpolation. Last, the final segmentation slices were resampled to the

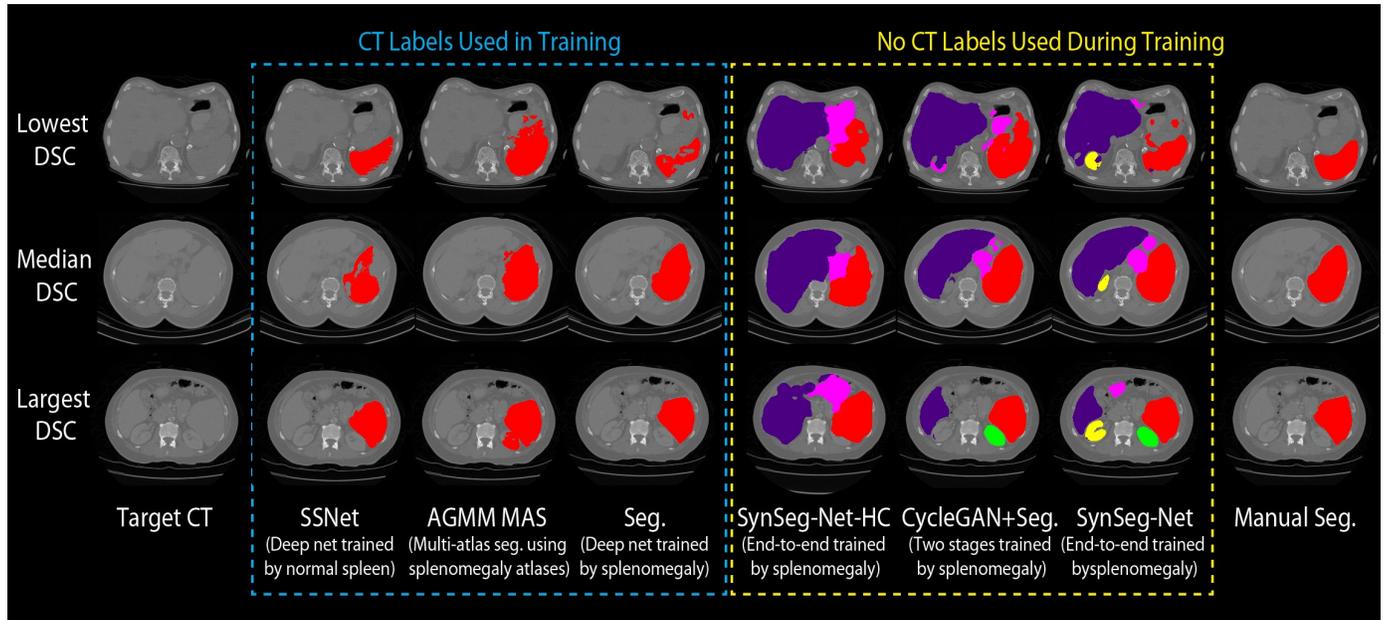

Fig. 5. The qualitative results were presented in this figure, including (1) three canonical methods using CT manual labels in CT segmentation, and (2) CycleGAN+Seg. and the proposed SynSeg-Net methods without using CT manual labels. The splenomegaly CT labels were only used in validation and excluded from training for (2). Moreover, later methods not only performed spleen segmentation but also estimated labels for other organs, which were not provided by canonical methods when such labels were not available on CT.

Table 1. Dice similarity score (DSC) and average surface distance (ASD) for CT splenomegaly testing images.

|  | SSNet | AGMM MAS | Seg. | SynSeg-Net-HC | CycleGAN+Seg. | SynSeg-Net |
|---|---|---|---|---|---|---|
| Median DSC | 0.679 | 0.912 | 0.911 | 0.628 | 0.880 | **0.919** |
| Mean±Std DSC | 0.630±0.269 | 0.861±0.101 | **0.911**±0.040 | 0.605±0.084 | 0.878±0.056 | 0.895±0.063 |
| Median ASD | 8.882 | 3.164 | **2.005** | 15.181 | 5.835 | 2.864 |
| Mean±Std ASD | 18.340±27.991 | 6.726±7.710 | **3.004**±2.797 | 14.383±4.521 | 5.600±3.619 | 3.898±3.397 |

* the unit for ASD related measurements is millimeter (mm).

original resolution using nearest neighbor interpolation and were concatenated. During training, the 2D slices were sampled randomly across all scans without forcing each batch to have only consecutive slices or only from the same subject.

The experiments were performed on an Ubuntu workstation, with NVIDIA Titan GPU (12 GB memory) and CUDA 8.0. The code of preprocessing and processing was implemented in MATLAB 2016a (www.mathworks.com), while the code of SynSeg-Net methods was implemented in Python 2.7 (www.python.org). For DCNN methods, the PyTorch 0.2 version (www.pytorch.org) was used to establish the network structures and perform training.

*D. Evaluation Metrics*

The Dice similarity coefficient (DSC) was employed to evaluate different approaches by comparing their segmentation results against the ground truth voxel-by-voxel. Differences between methods were evaluated by Wilcoxon signed rank test [38] with a significance threshold of $p<0.05$.

## IV. EXPERIMENTAL DESIGN AND RESULTS

We conducted experiments on two different applications to evaluate the relative effectiveness of different approaches. The first application is the MRI to CT splenomegaly synthetic segmentation. The second application is the CT to MRI TICV synthetic segmentation. In the first experiment, we first employed the target abdominal CT intensity images in the training (without using the manual labels), which would provide the best synthetic segmentation performance since the target intensity images were used in the synthesis learning. Then, we use an independent CT cohort for validation.

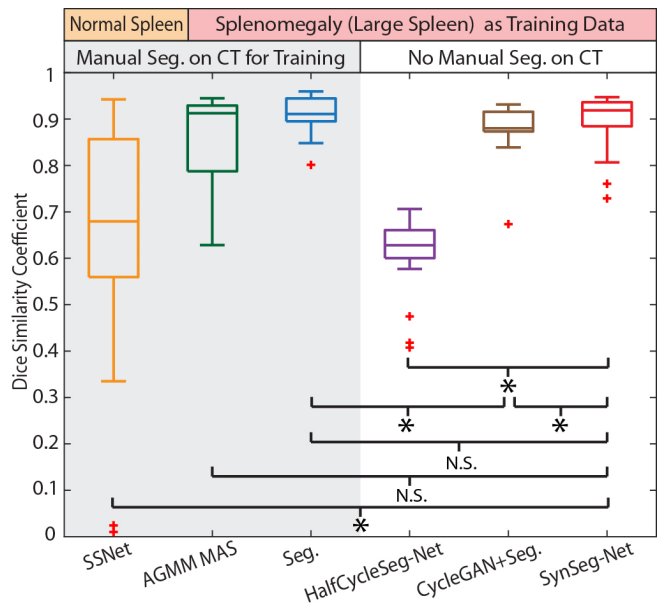

Fig. 6. The boxplot results of all CT splenomegaly testing images, where "*" means the difference are significant at $p<0.05$, while "N.S." means not significant.

### A. MRI-to-CT Splenomegaly Synthetic Segmentation for Abdomen

*1) Data*

A collection of 60 clinical acquired whole abdomen MRI T2w scans as well as 19 clinical acquired whole abdomen CT scans from splenomegaly patients were used as the training and testing data. The MRI and CT scans were acquired in the axial plane. In total, 3262 MRI slices and 1874 CT slices were used in the experiments.

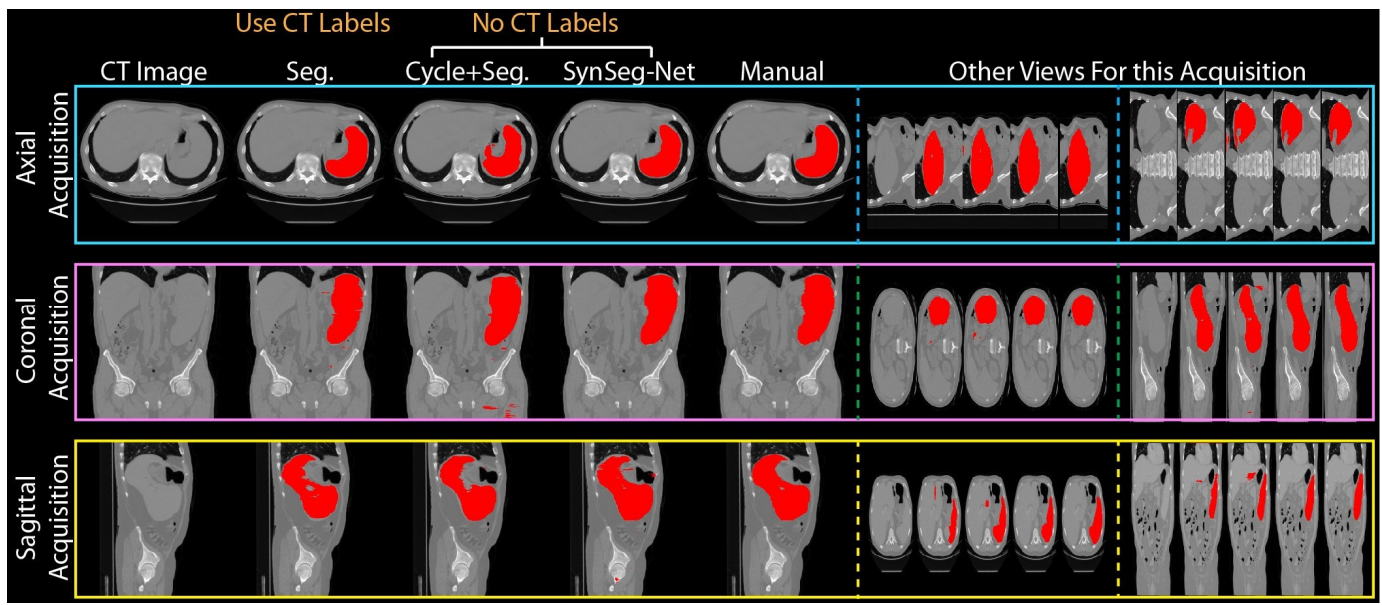

Fig. 7. The qualitative results were presented in this figure. Three rows indicated the three types of scans in the external validation cohort: (1) axial acquisition, (2) coronal acquisition, and (3) sagittal acquisition. The results of the corresponding acquisition view were presented in the left panels, while the results for remaining views were showed in the right panels.

## 2) Experimental Design
**CT Segmentation with CT Manual Labels**

First, our previously developed spleen segmentation network (SSNet) [31] (trained by 75 normal spleen CT scan) was employed to assess performance of a network trained by normal spleens applied to splenomegaly scans.

Then, multi-atlas segmentation and residual FCN network were used as two baseline methods, which used traditional segmentation strategies: trained by 19 splenomegaly CT scans as well as the corresponding manual spleen labels in a leave-one-out cross validation manner. Briefly, the adaptive Gaussian mixture model multi-atlas segmentation (AGMM MAS) was used as the first baseline method, which has been shown its superior performance on splenomegaly segmentation [39]. The second baseline approach employed the 9 block ResNet FCN [5, 36]. To compare with the synthetic segmentation methods, the network structure and the hyperparameters of the ResNet were kept exactly the same as the generators and segmentation networks in SynSeg-Net. This method evaluated the performance of traditional supervised DCNN segmentation, which used the manual labels in target imaging during training. Since only spleen manual labels were available in the CT domain, the supervised learning methods using CT manual labels provided spleen segmentation results.

**CT Segmentation without CT Manual Labels**

Then, we evaluated the performance of synthetic segmentation, which did not use the manual labels in target imaging modality during training. In this section, the two stage CycleGAN+Seg. strategy proposed by Chartsias et al. [35] as well as the proposed end-to-end SynSeg-Net were evaluated. To be a fair comparison, the network structures of CycleGAN+Seg. and SynSeg-Net were the same except that the SynSeg-Net, which connected the synthesis and segmentation in an end-to-end training. Briefly, the CycleGAN+Seg. strategy firstly trained the CycleGAN network to achieve 60 synthetic CT scans from 60 real MRI scans. Then the manual labels from real MRI scans as well as the corresponding synthetic CT scans were used to train an independent 9 block ResNet network. Hence, two independent training phrases were used.

By contrast, the proposed SynSeg-Net integrated the two synthesis and segmentation training phrases into an end-to-end training framework. The examples of real, synthesized, reconstructed and segmentation images for Path A and Path B were shown in Figure 4.

We also performed an experiment that trained the SynSeg-Net only using the source to target path (from MRI to CT) without using the target to source path. The experiment only used $G_1$ and $T$ in the half cycle (HC), which was called SynSeg-Net-HC. This experiment presented the segmentation performance with/without the complete cycle.

All networks were trained and validated for 100 epochs. The epoch with highest mean DSC between predicted and manual segmentation on 19 splenomegaly CT scans were reported in the results. The best performance of ResNet (epoch=90) was obtained from leave-one-subject-out validation. The best performance of SSNet (epoch=10), SynSeg-Net-HC (epoch=10), CycleGAN+Seg. (epoch=50) and SynSeg-Net (epoch=40) were evaluated from the external validation since labels for 19 splenomegaly CT scans were never used in the training. Since liver, left kidney, right kidney and stomach manual labels were avilable in additon to spleen labels in MRI, the corresponding automatic organ segmentation results were also presented qualitatively in Figure 5 for SysSeg-Net-HC, CycleGAN+Seg, and SynSeg-Net. However, we did not evaluate the results except spleen since (1) we did not have manual labels for the remaining organs in CT domain, (2) the purpose of this experiment is to perform spleen segmetnation.

## 3) Results

The qualitative and quantitative results were shown in Figure 5 and 6 respectively. Three subjects with largest, median and smallest DSC of SynSeg-Net were presented. From the results,

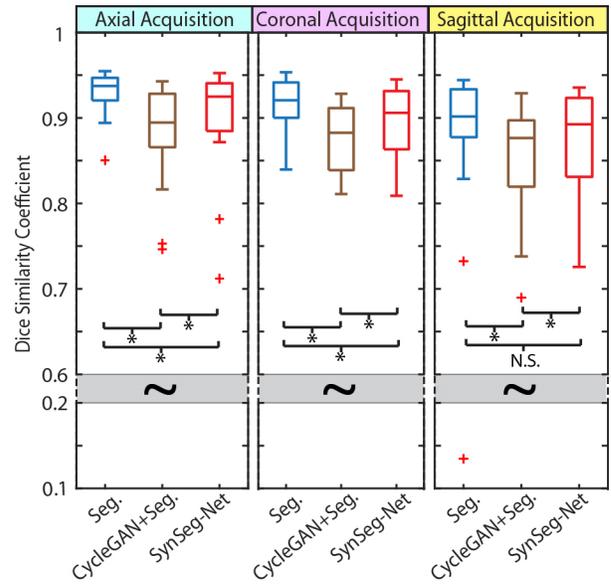

Fig. 8. The boxplot results of all CT splenomegaly external validation images, where "*" means the difference are significant at p<0.05, while "N.S." means not significant. Seg. was the ResNet segmentation network using manual labels in CT. Without using CT manual labels, CycleGAN+Seg. and SynSeg-Net were the two-stage and end-to-end network designs respectively.

Table 2. Dice similarity score (DSC) and average surface distance (ASD) for all CT splenomegaly external validation images.

|  | Seg. | Cycle+Seg. | SynSeg-Net |
|---|---|---|---|
| Axial Acquisition | | | |
| Median DSC | **0.937** | 0.894 | 0.925 |
| Mean±Std DSC | **0.929**±0.025 | 0.884±0.057 | 0.906±0.058 |
| Median ASD | **2.073** | 3.698 | 2.696 |
| Mean±Std ASD | **2.276**±1.005 | 4.074±2.216 | 3.157±1.804 |
| Coronal Acquisition | | | |
| Median DSC | **0.921** | 0.883 | 0.906 |
| Mean±Std DSC | **0.914**±0.031 | 0.876±0.037 | 0.882±0.087 |
| Median ASD | **1.840** | 3.508 | 2.125 |
| Mean±Std ASD | **2.116**±0.772 | 4.281±3.124 | 2.939±2.127 |
| Sagittal Acquisition | | | |
| Median DSC | **0.902** | 0.877 | 0.893 |
| Mean±Std DSC | 0.863±0.166 | 0.850±0.066 | **0.872**±0.064 |
| Median ASD | **2.308** | 3.863 | 2.932 |
| Mean±Std ASD | **3.934**±5.733 | 5.083±3.294 | 4.043±2.905 |

* the unit for ASD related measurements is millimeter (mm).

the SynSeg-Net was not only able to perform the spleen segmentation, but also estimated segmentations on liver, left kidney, right kidney and stomach. The "*" indicates the difference between methods were significant, while "N.S." means not significant. Average surface distance (ASD) measurements (median, mean, and standard deviation (Std)) were presented as well as the DSC measurements in Table 1.

Without using CT labels, the SynSeg-Net achieved significant superior performance compared with CycleGAN+Seg. and SynSeg-Net-HC methods, while achieving comparable performance with baseline ResNet segmentation network using CT labels.

### B. External Validation for MRI-to-CT Splenomegaly Synthetic Segmentation

In the previous experiment, the target images were used in the training stages (only intensity images were used and the labels were excluded). This strategy would provide the best performance of the proposed SynSeg-Net since the target images were used to model the target distributions. However, the training stage needs to be performed again for an unseen target image. A more general strategy is to apply the trained model on new target images directly. In this experiment, we employed an independent external validation cohort to evaluate the performance of the baseline segmentation network, two stages CycleGAN+Seg. network, and proposed SynSeg-Net.

#### 1) Data

A set of 66 whole abdomen CT scans from independent study were used as the external validation data to evaluate the performance of different method. Among the entire cohorts, 23 scans were axial acquisition, 21 scans were coronal acquisition, while 22 scans were sagittal acquisition (in Figure 7).

#### 2) Experimental Design

The same preprocessing was performed on the 66 CT scans before segmentation. Then, the trained baseline ResNet segmentation network (19 CT + CT labels), the trained Cycle+Seg. and SynSeg-Net (60 MRI and 19 CT + MRI labels) from the previous experiment were applied to segment 66 external validation CT scans directly. The parameters and the presented epochs were the same as the previous experiment without additional training or fine-tuning.

#### 3) Results

The qualitative and quantitative results have been showed in Figure 7 and 8 respectively. The "*" indicates the difference between methods were significant, while "N.S." means not significant.

In Figure 8, the boxplots were presented to compare the segmentation results. Without using CT labels, the SynSeg-Net achieved significant superior performance compared with CycleGAN+Seg. method, while achieved comparable performance with baseline ResNet segmentation network using CT labels for sagittal acquisition scans. The corresponding ASD measurements were presented with the DSC measurements in Table 2.

### C. CT-to-MRI TICV Synthetic Segmentation for Brain

The previous two experiments evaluated the performance of MRI to CT synthetic segmentation for an abdominal organ. Therefore, the synthesis was performed to learn a less context imaging modality (abdominal CT) from a richer context imaging modality (abdominal MRI). In this experiment, we evaluate the performance of CT to MRI synthetic segmentation for the brain, which is a more challenging task since brain MRI has much richer tissue context than brain CT. Therefore, this experiment evaluated the performance of synthetic segmentation on a richer context imaging modality (brain MRI), whose training images were synthesized from a less context imaging modality (brain CT).

#### 1) Data

20 subjects with both whole brain and MRI and CT were used in this experiment. True TICV labels were available for all 20 subjects, whose imaging parameters and the atlas generation were described in [40]. To evaluate the SynSeg-Net performance, we separate 20 subjects to two groups, group CT and group MRI. In group CT, the first half (10 subjects) were used, in which we supposed only CT scans were available. In group MRI, the remaining half (10 subjects) were used, in which we supposed only MRI scans were available. Therefore,

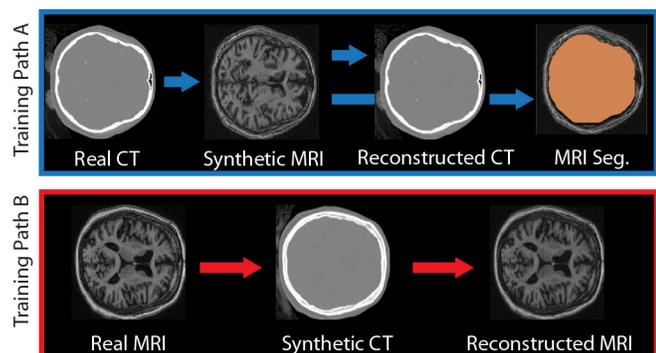

Fig. 9. The intermediate results of the real, synthesized, and reconstructed images as well as segmentations in training Path A and Path B.

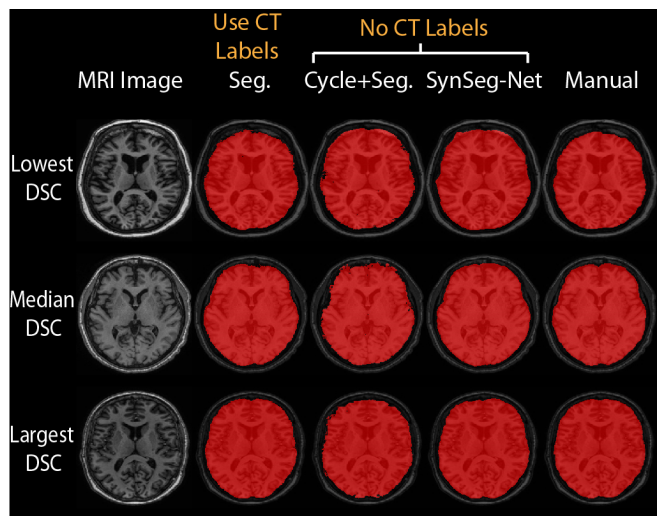

Fig. 10. The qualitative results were presented in this figure. Three rows indicated three subjects with largest, median and lowest DSC of SynSeg-Net were presented.

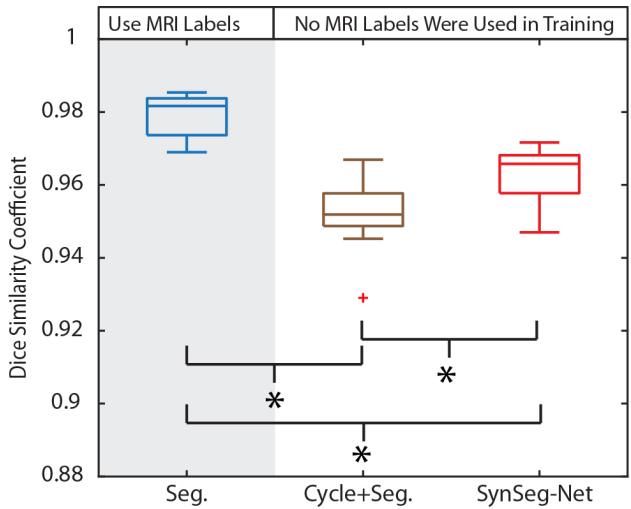

Fig. 11. The boxplot results of all MRI TICV testing images, where "*" means the difference are significant at p<0.05, while "N.S." means not significant. Seg. was the ResNet segmentation network using manual TICV labels in MRI. Without using MRI manual labels, CycleGAN+Seg. and SynSeg-Net were the two-stage and end-to-end network designs respectively.

Table 3. Dice similarity score (DSC) and average surface distance (ASD) for MRI TICV testing images.

|  | Seg. | Cycle+Seg. | SynSeg-Net |
|---|---|---|---|
| Median DSC | **0.982** | 0.952 | 0.966 |
| Mean±Std DSC | **0.979±0.006** | 0.952±0.010 | 0.963±0.008 |
| Median ASD | **0.803** | 2.118 | 1.312 |
| Mean±Std ASD | **0.987±0.458** | 2.322±0.976 | 1.441±0.318 |

* the unit for ASD related measurements is millimeter (mm).

we have two independent unpaired training data: 10 with CT as the source modality and 10 with MRI as the target modality. Since we aim to train an MRI TICV segmentation network, we only use the true TICV labels for 10 CT scans in the training. The true TICV labels for 10 MRI scans were excluded from training and only used during validation.

*2) Experimental Design*

Unpaired 10 CT scans and 10 MRI scans as well as the TICV true labels on CT were used to train the SynSeg-Net. After training, an MRI TICV segmentation network was achieved without using any TICV labels on MRI. The preprocessing steps are the same as the abdomen scans. The same baseline segmentation network, Cycle+Seg. network, and the proposed SynSeg-Net were employed in the validation. The hyperparameters for training all deep networks were kept the same as the experiments for abdomen scans. The examples of real, synthesized, reconstructed images and segmentation images for Path A and Path B were shown in Figure 9. Since the external validation data were not available for MRI TICV segmentation, all the methods were compared at epoch 200 consistently.
]
*3) Results*

The qualitative and quantitative results have been shown in Figure 10 and 11 respectively. The "*" indicates the difference between methods were significant, while "N.S." means not significant.

In Figure 11, the boxplots were presented to compare the segmentation results. Without using MRI TICV labels, the SynSeg-Net achieved significant superior performance compared with the CycleGAN+Seg. method, while yielded inferior performance compared with baseline ResNet segmentation network using MRI TICV labels. The corresponding ASD measurements were presented with the DSC measurements in Table 3.

## V. Conclusion and Discussion

SynSeg-Net enables training of a deep convolutional segmentation network without having ground truth labels in the target modality. Figure 6 showed that the SSNet trained by normal spleen CT images was significantly worse than other methods. The proposed SynSeg-Net method was significantly better than the two stages CycleGAN+Seg. method. Without using CT labels, the SynSeg-Net achieved the comparable performance as the AGMM MAS and ResNet that used CT labels. On the contrary, the performance of CycleGAN+Seg. was significantly worse than ResNet. Then, without including target intensity image during training, the proposed SynSeg-Net approach resulted in significantly superior performance compared with CycleGAN+Seg. method, while achieved comparable performance with baseline ResNet segmentation network using CT labels for sagittal acquisition scans (Figure 8). Last, for the CT to MRI TICV synthetic segmentation, the proposed SynSeg-Net approach resulted in significantly superior performance compared with CycleGAN+Seg. method (Figure 11).

One major limitation of deep learning based segmentation is the limited generalization ability across applications, domains and imaging modalities. Figure 6 showed that the SSNet trained on normal spleen did not provide decent segmentation performance on splenomegaly even when both were for the same organ within the same imaging modality. One solution is to label a set of images for the new application, however, the manual tracing is resource intensive. Therefore, it is appealing to reuse the previous manually labeled images from another modality as the proposed SynSeg-Net. Moreover, the images from source and target domains were not limited to paired ones. To be potentially compatible with other applications other than spleen segmentation, we employed the ResNet instead of SSNet as the generators and segmentation subnet since the ResNet is one of the most widely used network across a variety of medical image processing tasks. From Table 1, the proposed SynSeg-Net without using splenomegaly CT manual labels achieved superior performance compared with Cycle+Seg. method, while had comparable performance as ResNet using splenomegaly CT manual labels. From Table 3, although inferior than training directly using target modality manual labels, the performance of SynSeg-Net was better than Cycle+Seg. for TICV segmentation.

In SynSeg-Net, 9 block ResNet was used as the generators and the segmentation network as it was validated in the original CycleGAN paper [5]. Meanwhile, PatchGAN was used as the discriminators. While this combination is successful, we do not claim optimality of using ResNet or PatchGAN. Using other image to image generators, segmentation networks, or discriminators might yield better performance when performing synthetic segmentation. Since the proposed method is an open framework, users are encouraged to explore methods other than the components that used in this study. In the future, it would be also interesting to evaluate the inter-rater reliability by including the manual segmentations from different human raters. Such results will also provide the comparison between automatic methods and human experts.

In SynSeg-Net, a 2D network was used to perform synthesis and segmentation since the numbers of training images for both experiments were not large enough to train a reasonable 3D network. However, the proposed SynSeg-Net is able to be extended to 3D framework and we hypothesize that it would yield better 3D segmentation performance if large number of training scans are available. In fact, Zhang et al. [3] has showed the promising results when performing 3D synthesis and segmentation. The 2D slice wise nearest neighbor interpolation was used to resample a manual segmentation volume to be compatible with the network in this study. However, such step can be replaced by other resampling methods (e.g., registration-based method), which might yield better accuracy for larger numbers of output labels. During training, the 2D slices were sampled randomly across all scans. Therefore, the scans with more slices have more data during training, which would introduce bias due to the unbalanced sampling. Such bias can be alleviated by designing balanced sampling strategy (e.g., perform interpolations on raw scan) in the future.

The quantitative evaluation on "how good the synthesized images are?" is still an open question in our community, especially when the synthesized images were used as intermediate data for segmentation. A recent study even suggested that certain synthesized images should not be used for direct interpretation since the synthesized images may lead to misdiagnosis when transferring the distributions [41]. Therefore, we focus on developing a new strategy of image segmentation rather than providing a new image synthesis method. As a result, the quantitative evaluations in this paper were on the segmentation rather than on the synthesis. Moreover, the unpaired images were used in this study without using the multi-modal images from the same patient, which also limited the validation on synthesis. The proposed end-to-end framework might benefit the synthesis compared with the two-stage design. For instance, the intermediate results in the end-to-end framework might provide "better" intermediate representations for segmentation. Thus, the final segmentation performance was used as the metrics across different segmentation tasks to compare the proposed method with the two-stage method. In summary, the proposed end-to-end method achieved consistent superior segmentation performance compared with the two-stage design across different tasks.